\documentclass[runningheads]{llncs}
\usepackage{mathrsfs}
\usepackage{graphicx}
\begin{document}
%
\title{{Arianna$^{+}$}: Scalable Human Activity Recognition by Reasoning with a Network of Ontologies\thanks{Accepted in the 17th International Conference of the Italian Association for Artificial Intelligence - Trento, Italy, November 20-23, 2018.}}
\titlerunning{Scalable Activity Recognition by Reasoning with a Network of Ontologies}
%
\author{Syed Yusha Kareem\inst{1} \and
Luca Buoncompagni\inst{1} \and
Fulvio Mastrogiovanni\inst{1,2}}
\authorrunning{S. Y. Kareem et al.}
%
\institute{$^{1}$Department of Informatics, Bioengineering, Robotics and Systems Engineering, University of Genoa, Via Opera Pia 13, 16145, Genoa, Italy\\$^{2}$Teseo srl, Piazza Montano 2a, Genoa, Italy}
\maketitle              
%
\begin{abstract}
Aging population ratios are rising significantly.
Meanwhile, smart home based health monitoring services are evolving rapidly to become a viable alternative to traditional healthcare solutions.
Such services can augment qualitative analyses done by gerontologists with quantitative data.
Hence, the recognition of Activities of Daily Living (ADL) has become an active domain of research in recent times.
For a system to perform human activity recognition in a real-world environment, multiple requirements exist, such as scalability, robustness, ability to deal with uncertainty (e.g., missing sensor data), to operate with multi-occupants and to take into account their privacy and security.
This paper attempts to address the requirements of scalability and robustness, by describing a reasoning mechanism based on modular spatial and/or temporal context models as a network of ontologies. The reasoning mechanism has been implemented in a smart home system referred to as {Arianna$^{+}$}.
The paper presents and discusses a use case, and experiments are performed on a simulated dataset, to showcase {Arianna$^{+}$}'s modularity feature, internal working, and computational performance.
Results indicate scalability and robustness for human activity recognition processes.

\end{abstract}
%
%
\section{Introduction}
\label{sec:intro}
%
In recent times, there is a rise in population of elderly individuals, as it is estimated that approximately $20\%$ of the world's population will be age 60 or older by 2050~\cite{hp}.
This motivates the research community and technology companies to provide, at home, healthcare services for the elderly, such that they can live safely and independently for longer periods of time.
The ability to perform Activities of Daily Living (ADL) without assistance from other people can be considered as a reference for the estimation of the independent living level of the elderly individuals~\cite{Liu16}.
Nowadays, geriatrists judge the well being of elderly individuals by observing them while they perform ADL, such as \emph{walking} and \emph{dressing}.
When possible, they measure variations in both space and time domains, needed to perform particular ADL.
This is done in sessions at certain time intervals, e.g., each year, to make quantitative judgments. 
But for some ADL, e.g., \emph{eating}, they rely on qualitative judgments of how the activity is performed, based on questionnaires.
Instrumental ADL (IADL) are taken into account as well, with similar qualitative observations, since they require a certain level of planning capabilities and social skills, such as \emph{housekeeping}, \emph{cleaning}, and \emph{cooking}.

A quantitative assessment of such qualitative data can be provided by a smart home specialized for elderly care, as it can recognize activities performed throughout the day and report to geriatrists.
This would enable accurate health assessments based on continuous evaluations. As presented in~\cite{survey2018,brunoHAR}, depending on the kind of sensors employed in the smart home, activity recognition (AR) can be performed using data originating from vision, inertial, distributed sensors or a combination of them.
However, AR is enabled by \textit{a priori} AR modeling, and for this the approaches used in the literature are mostly of two types, one being data-driven, and the other being knowledge-driven.
A discriminative (e.g., Support Vector Machines and Artificial Neural Networks) type, data-driven approach is used when complex, multi-modal data streams are involved, e.g., data originating from cameras~\cite{cameraUse} and accelerometers~\cite{accelUse}, for posture recognition and fall detection.
When simpler data are involved (e.g., while using distributed sensors) either a generative (e.g., Hidden Markov Models and Dynamic Bayesian Networks) type, data-driven approach is taken~\cite{generative1}, or a knowledge-driven approach is adopted~\cite{Scalmato,luca}.
Although some sensors (e.g., cameras) provide high accuracy for monitoring individuals; due to privacy issues, simpler sensors (e.g., Passive Infrared (PIR), light, and Radio-Frequency Identification (RFID)) are largely used.

Learning (or development) of AR models, in data-driven approaches, happens by training over datasets, whereas in knowledge-driven approaches it is done by explicitly encoding knowledge, typically in the form of set of axioms, used for AR based on sensor data.
In terms of modularity with activity models, the former approach is not friendly since, if a new activity is to be introduced into the system, a new dataset has to be collected and the entire training process has to be performed. Whereas the latter approach is modularity friendly as a new activity model's knowledge can simply be added as a set of axioms and rules. 

In this paper, we describe a knowledge-based approach for domain modeling (i.e., of context/activity) and reasoning (i.e., context/activity recognition), which is currently part of our {Arianna$^{+}$} smart home framework.
The approach adopts: ($1$) Ontology Web Language (OWL), based on description logics (DL)~\cite{DL}, which is a fragment of first order predicate logic, designed to be as expressive as possible while retaining decidability. It allows to describe a given domain by defining relevant concepts (in the terminological box or TBox), and by asserting properties of individuals that are instances of those concepts (in the assertional box or ABox). Reasoners can then be used to derive facts, i.e., make implicit knowledge explicit, by reasoning mechanism~\cite{deduction1994} based on \textit{subsumption} of concepts and \textit{instance checking}. ($2$) Rules based on the Semantic Web Rule Language (SWRL)~\cite{swrl}, which allow the system to perform query and manipulations as a unique operation based on logic conjunctions.

Due to issues of language expressivity, OWL-DL reasoners do not perform temporal reasoning. Nevertheless, the idea of using OWL for AR can be found in the literature and~\cite{riboni} highlights that when ontological techniques are extended with even simple forms of temporal reasoning, their effectiveness increases.
Moreover, symbolic temporal concepts have been used for AR~\cite{chen}, and this is usually done using Allen's algebra~\cite{allen}, which allows DL reasoners to consider instances of time belonging to particular intervals.
In the literature, some attempts~\cite{Scalmato,luca} at ontology-based AR take temporal reasoning into account but accumulate temporal instances.
Hence,
their search space grows exponentially~\cite{uoftoroaodl} with respect to the number of axioms in the ontology, which is an issue for large-scale, real-time applications.
In this paper, we take basic temporal aspects for AR into account, without accumulating time instances within ontologies.

In a real-world environment, we argue that AR systems must carefully guarantee \emph{scalability} and \emph{robustness} requirements.
On the one hand, scalability can be achieved when
(i) the system is \emph{modular} with respect to activity models and
(ii) types of sensors, as well as, 
(iii) is able to manage computational resources and memory, since they affect recognition performance in long-term applications~\cite{luca2017forgettings}.
On the other hand, robustness, which is a more strict requirement to be achieved, strongly depends on the \textit{design} of the activity models.
We also argue that a redundancy of models, with which we can assess the same activity, can increase the overall system's robustness. 
The above-listed requirements lead respectively, to the issues of:
(i) designing modular activity models as part of an ontology network, which is able to infer activities based on the occurrence of events,
(ii) designing a system's architecture that incorporates distributed sensors data, and
(iii) designing the activity models such that they represent the context over time, and evaluate them with the most suitable behavior (e.g., with a scheduled frequency).

This paper extends the work presented in~\cite{luca}, wherein we propose to use a hierarchy of ontologies, that decouple logic operations for semantically describing the context and support modular composition of reasoning behaviors for online activity recognition.
Here, we present an AR-enabled smart home system from a software architecture perspective, and an implementation of a relevant use case, which is tested based on simulated data from distributed sensors. Furthermore, we address the issues presented above and highlight the modularity features and performance of {Arianna$^+$}, while reasoning over an ontology network. 

The paper is organized as follows.
Section \ref{sec:activity_detection} discusses the modular ontology network.
{Arianna$^+$}'s architecture is presented in Section \ref{sec:systems_architecture}, whereas Section \ref{sec:use_case_setup} discusses an implementation of a use case.
Finally, conclusions follow.
%
%
\section{Activity Detection}
\label{sec:activity_detection} 
\subsection{Dynamic Ontology Networks}
\label{sec:dynamic_ontology_network}
%
In~\cite{luca}, an ontology network is defined as a graph $G$,
wherein the set of \emph{nodes} $N$ are ontologies (each with an independent DL reasoner) containing \emph{statements} of the form~(\ref{eq:statement}), i.e., having a Boolean state $s$ and a generation timestamp $t$: 
\begin{equation}
\texttt{Statement} \sqsubseteq =_1 \texttt{hasState}(s) \sqcap =_1 \texttt{hasTime}(t)
\label{eq:statement}
\end{equation}
and are used to describe a specific part of the \textit{context}, while the set of directed \emph{edges} $E$ are communication channels used for sharing statements between the nodes. Hence $G$ is of the form:
\begin{equation}
G = \{N,E\}
\label{eq:graphGeneric}
\end{equation}
where, $N = n_1,n_2,\ldots,n_n$, such that each node specializes in reasoning within a particular context, and $E = e_{12},e_{13},\ldots,e_{1n},e_{21},e_{23},\ldots,e_{2n},\ldots,e_{mn}$, such that the index of each edge signifies the direction of flow of statements, e.g., in $e_{12}$ statements flow from $n_1$ to $n_2$.
Consider an \emph{event}, indicating that water is flowing from the sink in the kitchen. It can have different interpretations for a system aimed at recognizing activities such as cooking or cleaning.
Instead of recognizing them actively from the same representation, with an ontology network it is possible to decouple their models in order to reason upon them based on an event or set of occurring events. Where, an event occurs based on rules that aggregate statements by logical conjunction.
We show in the following Sections that this approach enforces system's modularity with respect to activity models, and if the network is such that it evaluates only the models related to a specific part of the overall context, then it also decreases the computation time. 

The system checks the statements in the network with a given frequency and, when an event is detected, specific external \emph{procedures} are executed in order to: (i) \textit{move} statements from one node to another via edges, and
(ii) evaluate models for activity recognition. 
For instance, statements could be generated from distributed sensors (e.g., detecting that Adam is in the kitchen at 8:00 am), then the system aggregates this information with prior knowledge to detect events (e.g., Adam is in the kitchen in the morning).
When such an event occurs, the model for detecting that Adam is having breakfast gets evaluated by checking statements and their temporal relations within the model.

Moreover, activity models can generate statements, e.g., indicating that Adam had (or did not have) breakfast at a certain time, and hence can trigger new events, which can further be used to describe the context and evaluate models via procedure executions.
A formal algebra of statements, used for defining events that execute procedures based on the context, has been proposed in~\cite{luca}.

\subsection{A Network of Activity Detectors}
\label{sec:a_network_of_activity_detectors}
\begin{figure}[t!]
\centering
\includegraphics[width=1\linewidth]{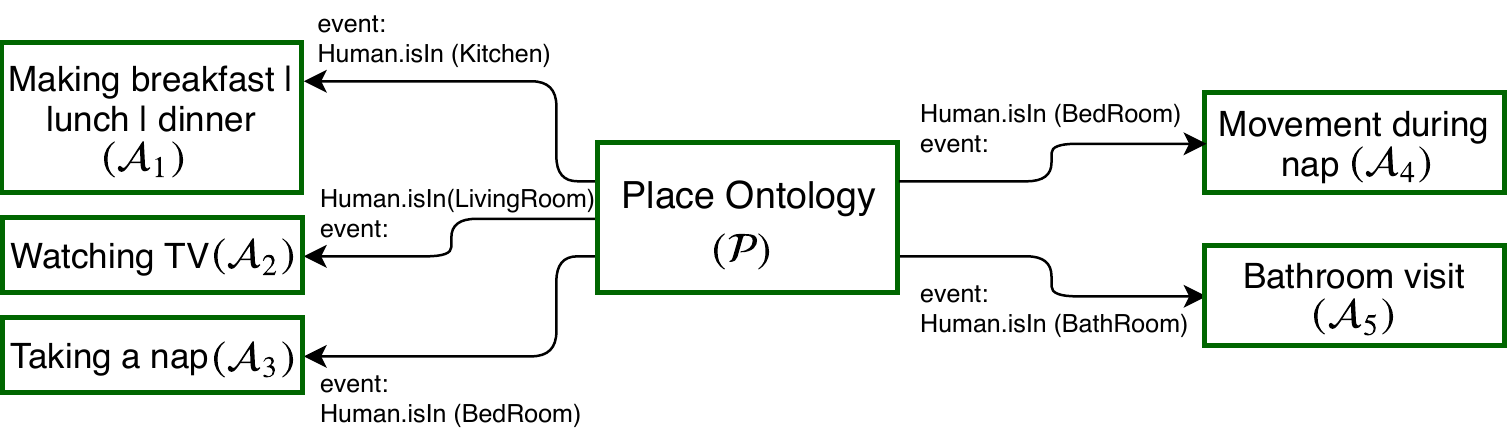}
\caption{A simplified ontology network $\mathcal{O}$.}
\label{fig:POAO}
\end{figure}
%
For the sake of description, we consider a simplified ontology network $\mathcal{O}$ as shown in Figure~\ref{fig:POAO}. In it there are $6$ nodes; $n_1$ is a location-based contextualizing model called \textit{Place Ontology} $\mathcal{P}$ and $n_2,\ldots,n_6$ are called \textit{activity models} $\mathcal{A}_i$, where $i=1,\ldots, 5$ respectively. Nodes are designed such that statements within $\mathcal{P}$ take into account the spatial aspect, and statements within $\mathcal{A}_i$ take into account the spatial and temporal aspects of AR.
$\mathcal{A}_i$ are listening for particular events that $\mathcal{P}$ generates, and the edges that link them are the following $E = e_{12},e_{13},e_{14},e_{15},e_{16}$. The nodes communicate and statements flow between them via edges, such that, $\mathcal{A}_i$ get activated and then evaluated by their independent reasoners, when a particular event occurs, as depicted by the graph in Figure~\ref{fig:POAO}.
If the evaluation of an activity model gets satisfied, its procedure generates a new statement to notify the recognition of an activity, e.g., \texttt{WatchingTV.\{hasState(True), hasTime(19:28)\}}.
\begin{figure}[t!]
\centering
\includegraphics[width=0.5\linewidth]{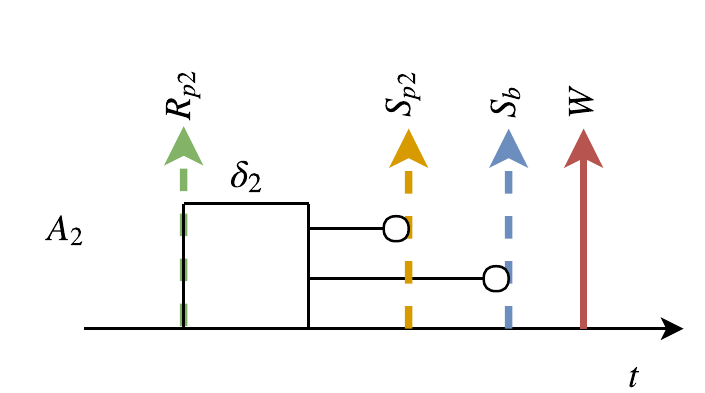}
\caption{Visual representation of statements that make up the $\mathcal{A}_2$ model: statements are shown as vertical arrows where dashed arrows indicate information from $\mathcal{P}$, and solid arrows indicate statements generated by this model. Statement indexes indicate sensors influencing the state of that statement, while the temporal restrictions are shown as black lines.}
\label{fig:swrlAO2}
\end{figure}
%

Within activity models, particular statements and temporal relations, must get satisfied for successful activity recognition.
These are shown for $\mathcal{A}_2$, which recognizes the activity \emph{WatchingTV}, in Figure~\ref{fig:swrlAO2}.
In it, statements are vertical arrows pointing upwards to indicate a $\emph{True}$ state and downwards for $\emph{False}$. 
These statements are either transferred from another node (e.g., dashed arrows represent statements coming from $\mathcal{P}$), or are generated by this node (e.g., solid arrows are the statements generated by $\mathcal{A}_2$) and are indicated along with a name and an index or a range of indexes.
A name is denoted by a capital letter and the sensors related to it are shown as the index.
Statements are annotated along a relative $x$-axis, in order to restrict their temporal relations through black lines ending with a circle.
In the Figure, we can see $4$ statements:
(i) statement $R_{p2}$, which is a dashed arrow of green color, is information coming from $\mathcal{P}$; it signifies \texttt{isIn\_LivingRoom.\{hasState(True), hasTime(19:25)\}}, where the index $p2$ indicates that the sensor \texttt{PIR2} influences the state of this statement;
(ii) statement $S_{p2}$, which is a dashed arrow of orange color, is information coming from $\mathcal{P}$; it signifies that there is some motion in the living room after $\delta_2$ time units, naively representing the idea that, if Adam is sitting on the sofa then he is not sitting still;
this statement can be replaced by a much robust statement, for instance, \texttt{sitting.\{hasState(True), hasTime(19:26)\}}, given that there may be other sensors in the system (e.g., wearable sensors, pressure sensors in the sofa);
(iii) statement $S_b$, which is a dashed arrow of blue color, is information coming from $\mathcal{P}$;
it signifies \texttt{highBrightnessTV.\{hasState(True), hasTime(19:28)\}}, where the indexes $b$ indicates that \texttt{brightness sensor} influences the state of this statement;
(iv) statement $W$, which is a solid arrow of red color, is generated when the overall model is satisfied, it signifies \texttt{WatchingTV.\{hasState(True), hasTime(19:28)\}};
this happens when statements $S_{p2}$ and $S_b$ are generated after $\delta_2$ time units with respect to the $R_{p2}$ statement.

With respect to the AR system presented in~\cite{luca}, the difference in the implementation of {Arianna$^+$} is two-fold. 
Firstly, in~\cite{luca} time-related instances get accumulated in the models for the purpose of temporal reasoning, and after an activity is recognized, the statements are removed to reduce the increasing complexity of the ontologies. 
In {Arianna$^+$}, when $\mathcal{A}_i$ receive statements from $\mathcal{P}$ the values of old instances get updated, if they are available.
This has the effect of not accumulating statements in $\mathcal{A}_i$, i.e, the procedure related to it is in charge of updating and evaluating it, without accumulating time-related instances.
Such a procedure performs temporal reasoning using both symbolic relations (inferred by the DL reasoner) and numerical/logical operations on the timestamps (inferred externally).
This approach of using an external reasoner has the affect of overcoming DL limitation, such as the issue of finding the minimum value in a set of numbers under the open world assumption. 
Secondly, events are queries that return Boolean value when certain statements are satisfied, or not, in an ontology of the network.
In~\cite{luca} events are semantically defined in an upper-ontology that schedules related procedures if their query is verified. Whereas in {Arianna$^+$} rather than having an upper-ontology, we have designed a system's architecture that incorporates the object-oriented programming (OOP) paradigm to execute $\mathcal{A}_i$ procedures with an \textit{event-listener} pattern.
%
%
\section{{Arianna$^+$}'s Architecture}
\label{sec:systems_architecture} 
\subsection{From Sensing to Context Awareness}
\label{sec:from_sensing_to_reasoning_layers} 
\begin{figure}[t!]
\centering
\includegraphics[width=1\linewidth]{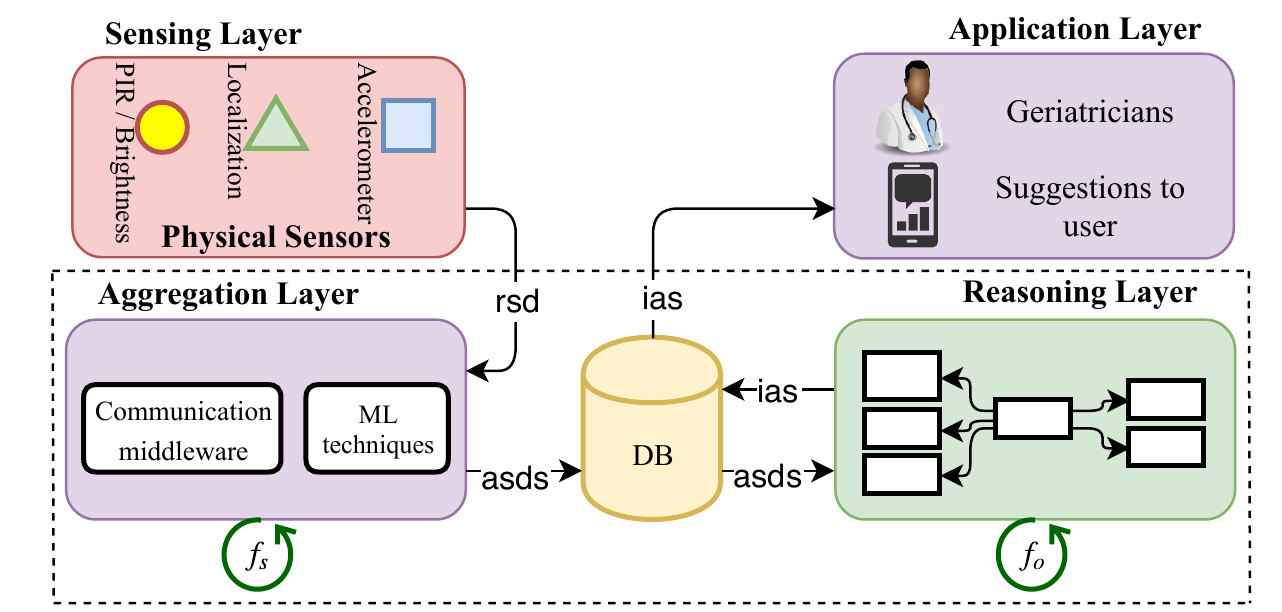}
\caption{{Arianna$^+$}'s architecture where the link \emph{rsd} signifies the flow of raw sensor data, \emph{asds} signifies the flow of aggregated sensor data in the form of statements, \emph{ias} signifies inferred activity statements and $f$ signifies frequency.}
\label{fig:sysarch}
\end{figure}

Figure~\ref{fig:sysarch} shows the system's architecture. It recognizes activities with $\mathcal{O}$ as described above; it comprises of the \emph{sensing}, \emph{aggregation}, \emph{reasoning} and \emph{application} layers.
In this Section, we focus on the interfaces between those layers, which enable the modular features of {Arianna$^+$} as highlighted in Section~\ref{sec:activity_detection}.
Firstly, in the reasoning layer, $\mathcal{O}$ is used over time for recognizing activities based on data taken from the database (DB), which is getting accumulated with the latest sensor values and timestamps by the aggregation layer, which in turn is connected to the physical sensory layer.
Finally, the application layer is used to easily interface geriatricians, other medical staff, assisted people and their relatives with {Arianna$^+$}'s services.

The \emph{reasoning layer} is {Arianna$^+$}'s core.
It is made up of $\mathcal{O}$ and its internal working is as described in Section~\ref{sec:a_network_of_activity_detectors}.
There are two components in the working of this layer.
The first is the initialization of $\mathcal{O}$ (i.e, TBox of ontologies are defined as nodes. While procedures and events are defined as edges).
The second is the frequency $f_o$ with which, in $\mathcal{O}$, the procedure of $\mathcal{P}$ takes in aggregated sensor data statements (link \emph{asds}) from the database, updates the ABox, reasons (spatially) with knowledge within $\mathcal{P}$, and declares occurrence of an event, if any.
If the declared event is being listened for by one or many $\mathcal{A}_i$, then their procedures get activated.
Once an activity model's procedure is active, it takes in statements from $\mathcal{P}$ and updates its own ABox, then reasons (spatially and temporally) with knowledge within the model and declares the recognition of a user activity.
This completes a chain of reasoning processes (i.e., $\mathcal{P}$ \emph{plus} an activity model), and if an activity is recognized in the process, then the procedure associated with the model saves the inferred activity statement (link \emph{ias}) back in the database.
As the reasoning process has not negligible computational time, if it is simply performed every time new sensor data statements arrive in the database, and if the frequency with which the new data arrives is faster than the reasoning process, then the system would not meet the near real-time constraint.
Hence, we need $f_o$ to have control over such a process.
It deals with the computational complexity issue of the DL reasoner which performs the reasoning in $\mathcal{O}$.

From the \emph{application layer}, on the one hand, geriatricians could visualize statistics related to the activities performed and explore further details in terms of statements (link \emph{ias}), if necessary.
On the other hand, the elderly individual could be stimulated with suggestions based on activity recognition, for instance, through dialogue-based interfaces via virtual coaches. 
Furthermore, the database also contains detailed logs of statements that were in $\mathcal{O}$, and therefore assistive or medical staff can access those statements to provide online services to the assisted individuals.
For instance, a future scenario of in-home healthcare would be such that, if Adam is asked by his doctor about the number of times he visits the bathroom during the night, Adam's reply can be augmented by quantitative data from the smart home, which can help the doctor in making healthcare-related decisions.

The \emph{aggregation layer} takes raw sensor data (link \emph{rsd}) from heterogeneous sensors in the \emph{sensing layer} and by using dedicated perception modules, processes the raw data to generate statements of the form (\ref{eq:statement}).
Then, it stores aggregated sensor data statements (link \emph{asds}) in the database.
This layer relies on a communication middleware module to channel all the Boolean data the sensors generate, and stores them in the database, if simple distributed sensors are considered.
Furthermore, it relies on classification modules (e.g., obtained via machine learning approaches) that can provide statements with semantics (e.g., \textit{sitting down}, \textit{lying down}, etc), and stores them in the database, i.e, if sensors generating more complex data streams are considered.
Remarkably, having a formal structure for a statement not only assures a modular evaluation of activity models, but also enables the overall AR system to take heterogeneous sensors into account.
Statements are stored in the database at a frequency $f_s$, and moreover each perception module in this layer can have its own frequency at which it processes the raw sensor data to generate statements and store them in the database.

It is noteworthy that the frequencies $f_s$ and $f_o$ are independent of each other, such that,
(i) the aggregation layer stores latest aggregated sensor data statements in the database at a frequency $f_s$, which can be unique for different perception modules, and (ii) the reasoning layer reasons based on the latest statements that are available to it from the database, with a frequency $f_o$.
%
%
\section{Use Case Setup}
\label{sec:use_case_setup} 
\subsection{Activity Models and Simulation Setup}
\label{sec:activity_models_and_simulation_setup} 

The use case considered in this paper utilizes all $\mathcal{A}_i$ in $\mathcal{O}$, as shown in~Figure~\ref{fig:POAO}.
Their description is as follows.
$\mathcal{A}_{1}$ infers \emph{Making breakfast, lunch or dinner}.
It is listening for the event \texttt{$\exists$ Human.isIn(Kitchen)}.
It generates one of the statements, \texttt{Making breakfast} or \texttt{Making lunch} or \texttt{Making dinner}, when the assisted person uses furniture (e.g., the kitchen cabinet), after being present in the kitchen for a minimum time period of $60$ seconds, and if that time period is inside one of the \textit{a priori} defined intervals of the day, i.e., morning, afternoon or evening.
$\mathcal{A}_{2}$ infers \emph{Watching TV}. It is listening for the event \texttt{$\exists$ Human.isIn(LivingRoom)}. 
It generates the statement \texttt{Watching TV} when the occupant uses furniture (e.g., the TV), after being present in the living room for a minimum time period of $60$ seconds, during any time of the day.
$\mathcal{A}_{3}$ infers \emph{Taking a nap in morning, afternoon or evening}.
It is listening for the event \texttt{$\exists$ Human.isIn(BedRoom)}.
It generates one of the statements \texttt{Taking a nap in morning} or \texttt{Taking a nap in afternoon} or \texttt{Taking a nap in evening}, when the assisted person uses furniture (e.g., the bed), after being present in the bedroom for a minimum time period of $60$ seconds, and if that time period is inside one of the intervals of the day, i.e., morning, afternoon or evening.
$\mathcal{A}_{4}$ infers \emph{Movement during nap}.
It is listening for the event \texttt{$\exists$ Human.isIn(BedRoom)}.
It generates the statement \texttt{Movement during nap} when the person uses furniture (e.g., the bed) and the PIR associated with the bed remains active even after $60$ seconds have passed on the bed, during any time of the day.
$\mathcal{A}_{5}$ infers \emph{Bathroom visit in morning, afternoon, evening or night}.
It is listening for the event \texttt{$\exists$ Human.isIn(BathRoom)}.
It generates one of the statements \texttt{Bathroom visit in morning} or \texttt{Bathroom visit in afternoon} or \texttt{Bathroom visit in evening} or \texttt{Bathroom visit in night}, when the assisted person uses furniture (e.g., the toilet seat), after being present in the bathroom for a minimum time period of $60$ seconds, and if that time period is inside one of the intervals of the day, i.e., morning, afternoon, evening or night.
\begin{figure}[t!]
\centering
\includegraphics[width=0.9\linewidth]{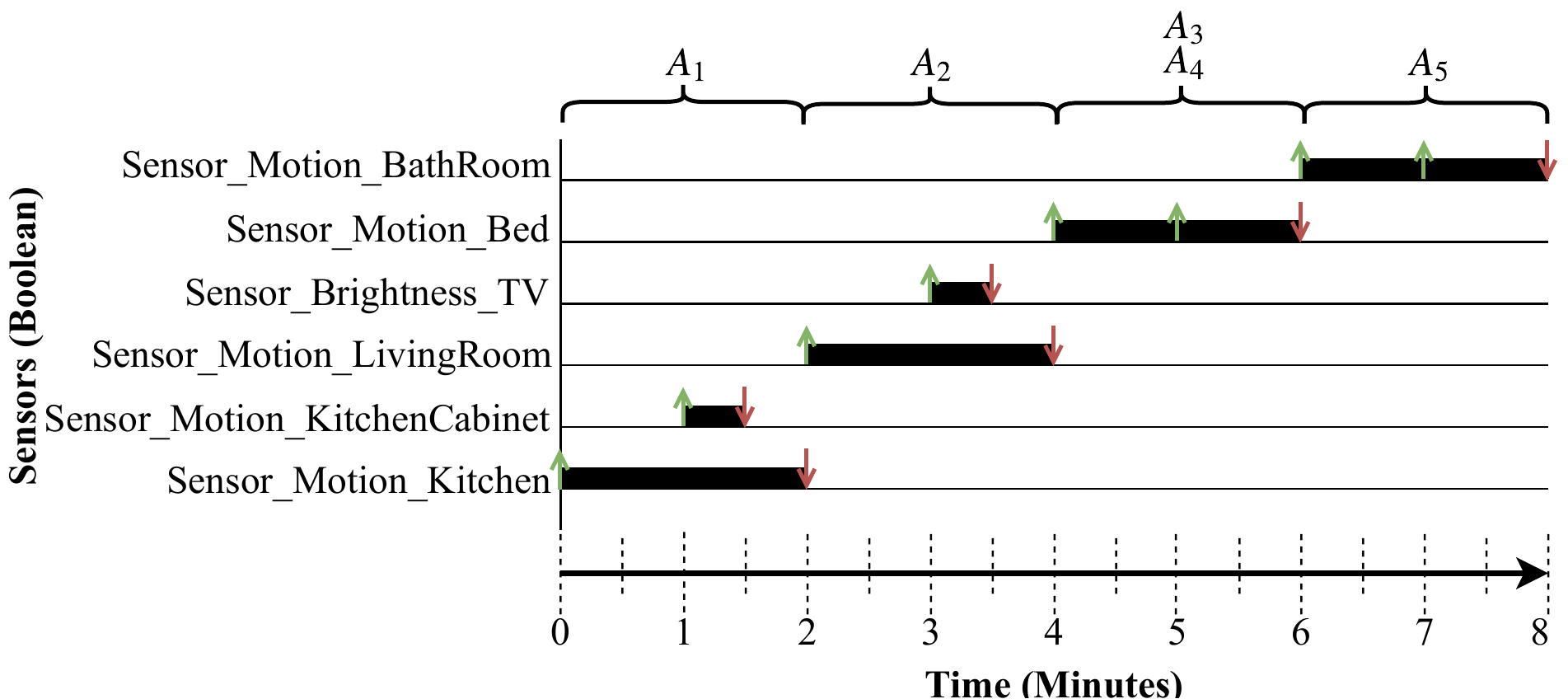}
\caption{The simulated dataset used in implementation of the use case.}
\label{fig:dataset}
\end{figure}

The use case is implemented by generating a simulated dataset with values and timestamps of a set of PIR sensors and a brightness sensor.
It depicts a scenario where an assisted person performs stereotypical activities that are held for eight minutes. 
The dataset is kept small so as to do extensive in-depth performance testing.
The simulation is performed by updating the database with simulated sensor data in the form of statements (mimicking the link \emph{asds} connecting the aggregation layer and the database).
As shown in Figure~\ref{fig:dataset}, Adam enters the kitchen, and after spending a minute in the kitchen, he opens the door of the kitchen cabinet and then closes it. 
He is in the kitchen for a total duration of 2 minutes. 
Next, he goes to the living room. 
After a minute in the living room, he switches on the TV and then switches it off after $30$ seconds. 
He is in the living room for a total duration of 2 minutes. 
Next, he goes to the bedroom and simulates sleeping on the bed. 
He does not stay still in the bed, rather is constantly in motion. 
He is in the bedroom for a total duration of 2 minutes. 
Finally, the person goes to the bathroom. 
He is in the bathroom for a total duration of 2 minutes. 

Among open source ontology reasoners that exist, e.g., Fact++, Pellet, Hermit and ELK. We use Pellet as it has more features in comparison~\cite{reasonersSurvey} and is able to pinpoint the root contradiction or clash when inconsistency occurs. Experiments have been performed on a workstation with the following configuration: Intel\textsuperscript{\tiny\textregistered} Core{\tiny\textsuperscript{\texttrademark}} $i7$ $2.6$ GHz processor and $8$ GB of memory. 
For assessing the system's performance, two types of evaluations are performed and compared.
The first is the Contextualized Activity Evaluation (CAE), and the second is the Parallel Activity Evaluation (PAE).
The CAE case represents the working of $\mathcal{O}$ as described in Section~\ref{sec:a_network_of_activity_detectors}, where $\mathcal{P}$ behaves as a contextualizer such that an activity model gets activated based on the context.
In the PAE case, $\mathcal{P}$ is no longer made to behave as a contextualizer, hence $\mathcal{A}_i$ are active in all contexts.

An evaluation (CAE or PAE) is performed as an experiment by setting a particular frequency $f_o$ (of the reasoning layer).
An experiment is performed with $5$ iterations, with each iteration an extra activity model is added to $\mathcal{O}$ to increase the system's complexity.
Each iteration is repeated $10$ times to assess the reasoner's average computational time, and the maximum and minimum variance, from among $10$ values.
In total four experiments are performed, their process and results are described in the following Section. 
\subsection{Performance Assessment}
\label{sec:performance_assessment}
\begin{figure}[t!]
\centering
\includegraphics[width=1\linewidth]{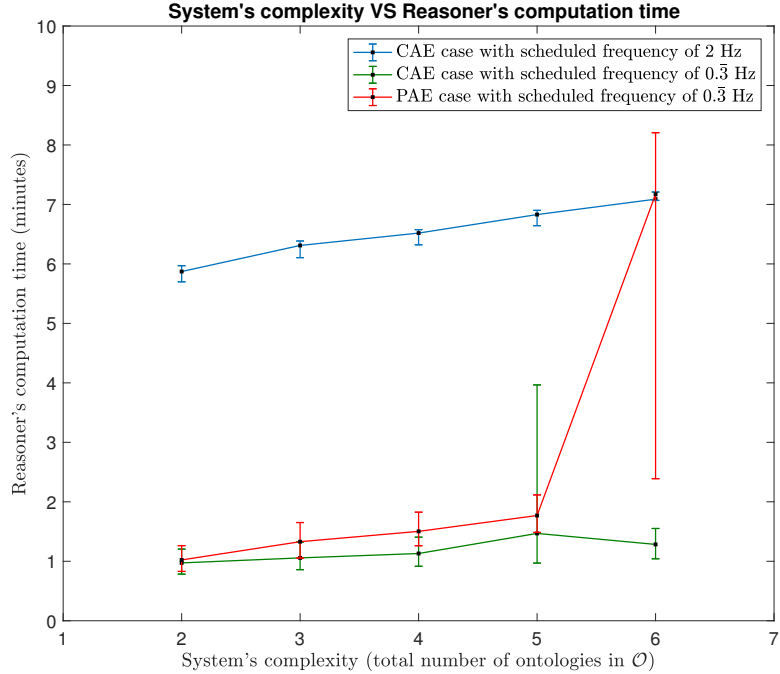}
\caption{System's complexity \textit{versus} reasoner's computation time. On $x$-axis are the number of ontologies, where $2$ means ($\mathcal{P}+\mathcal{A}_1$), $3$ means ($\mathcal{P}+\mathcal{A}_1+\mathcal{A}_2$), etc.}
\label{fig:results}
\end{figure}
%
Performance results are shown in Figure~\ref{fig:results}, where $x$-axis shows the increasing number of ontologies in $\mathcal{O}$ (i.e., the number of activities {Arianna$^+$} attempts to recognize), with each iteration of an experiment.
In relation to this, the $y$-axis shows the reasoner's computational time (i.e, the sum of the reasoning time spent in the ontologies of $\mathcal{O}$).
A thread with a unique color represents a unique experiment conducted with a particular scheduled frequency $f_o$.
A black dot on a thread marks the reasoner's average ($10$ repetitions of an iteration for an experiment) computational time, and vertical lines in the positive/negative direction (from a black dot) show the maximum/minimum variance, respectively, from the average computational time.
The simplest network has two ontologies, the Place Ontology $\mathcal{P}$ and $\mathcal{A}_1$, while the most complex network we tested has six ontologies, i.e., $\mathcal{P}$ and $\mathcal{A}_1, \mathcal{A}_2, \ldots, \mathcal{A}_5$.

Considering the CAE case, the reasoning layer is set to run with a time period of $500$ milliseconds (i.e., $\mathit{f}_o$ is $2$ Hz), with the hypothesis that recognizing activities within $500$ milliseconds is satisfying soft real-time constraint.
Represented by the blue thread, the reasoner's computational time is high and increases linearly with the increase in system's complexity.
Following the success of the previous test, and considering the PAE case, $\mathit{f}_o$ is kept the same, i.e, $2$ Hz.
However, this case is not represented by any thread, as an undefined amount of time was being taken by the reasoner to finish the reasoning process. 
Following the drawback in the previous case, and considering the same case, i.e., PAE, the reasoning layer is set to run with a higher time period of $3000$ milliseconds (i.e, $\mathit{f}_o$ is $0.\bar{3}$ Hz), to make sure that the reasoning process completes within the frequency $\mathit{f}_o$, a condition which is satisfied with that time period.
Represented by the red thread, the reasoner's computational time is initially low but then behaves exponentially, with the increase in system's complexity.
Finally, following the success of the previous case, and considering the CAE case, $\mathit{f}_o$ is kept the same, i.e, $0.\bar{3}$ Hz.
Represented by the green thread, the reasoner's computational time is initially low and remains low, as it increases linearly with the increase in complexity of the system.

More in the discussion of the results:
\begin{enumerate}
\item Comparing the two CAE cases with frequencies $2$ and $0.\bar{3}$ Hz, respectively, against each other, and against the PAE case with frequency at $2$ Hz, we see that, with the approach described in Section~\ref{sec:a_network_of_activity_detectors} (i.e., represented by the CAE cases), it is possible to have activity recognition with a high frequency. 
This shows {Arianna$^+$}'s ability to serve near real-time applications. 
\item In case of PAE, when there are $6$ ontologies in $\mathcal{O}$, a high variance is seen with the reasoner's average computational time on the higher end, thus confirming an exponential behavior.
In case of CAE when there are $5$ ontologies in $\mathcal{O}$, a high variance is seen with the reasoner's average computational time on the lower end, thus confirming a linear behavior.
\item The PAE case (wherein, even if multiple smaller ontologies are used, all their reasoners are running in parallel), represented by the red thread; shows an evident exponential behavior and can be compared to using one large ontology in the system.
As We know from the literature that (see Section~\ref{sec:intro}), with an increase in the number of axioms in an ontology the search space increases exponentially. 
Therefore in comparison, the CAE case, with its linear behavior, shows clearly the advantage of {Arianna$^+$}'s modularity feature with respect to activity models and their contextualized evaluation.
Furthermore, such claim is supported by the fact that our system does not accumulate instances within ontologies as it uses an external reasoner to deal with temporal aspects of reasoning (as described in Section~\ref{sec:a_network_of_activity_detectors}) and stores the recognized activities in a database (as mentioned in Section~\ref{sec:systems_architecture}).
%
%
%
\end{enumerate}
%
%
\section{Conclusion}
\label{sec:conclusion} 
%
In this paper, we present the activity recognition structure of our smart home framework {Arianna$^+$}, whose core is a reasoning layer based on an ontology network, which is grounded on ontology models, statements, procedures, and events-listeners, for which we provide general-purpose definitions. 
A use case scenario comprising of $5$ activity models was implemented and experimentally evaluated for assessing its behavior and computational performance. Results (with CAE) indicate that an AR system which exploits the modularity feature of a network of ontologies in a contextualized manner, and in which temporal instances are not accumulated, has near real-time AR capability and it addresses the scalability and robustness requirement.
Limitations of the presented use case are that it considers a single occupant in the environment and although extensively, it is tested with a simulated dataset. Hence, future work involves testing with data from a real distributed sensor scenario and incorporating perception modules in the aggregation layer, such that the network of ontologies can take statements related to human gestures and postures. 
Nevertheless, this paper provides a general-purpose discussion about ontology networks for AR. While the full evaluation of this approach awaits further investigation and user feedback, our initial results provide a base for building real-world use cases. 
%
%
%
%
\bibliographystyle{splncs04}
\bibliography{MyBib.bib}

\begin{thebibliography}{10}
\providecommand{\url}[1]{\texttt{#1}}
\providecommand{\urlprefix}{URL }
\providecommand{\doi}[1]{https://doi.org/#1}

\bibitem{reasonersSurvey}
Abburu, S.: A survey on ontology reasoners and comparison. International
  Journal of Computer Applications  \textbf{57}(17) (2012)

\bibitem{allen}
Allen, J.F.: Maintaining knowledge about temporal intervals. In: Readings in
  Qualitative Reasoning about Physical Systems, pp. 361--372. Elsevier (1990)

\bibitem{accelUse}
Atallah, L., Lo, B., Ali, R., King, R., Yang, G.Z.: Real-time activity
  classification using ambient and wearable sensors. IEEE Transactions on
  Information Technology in Biomedicine  \textbf{13}(6),  1031--1039 (2009)

\bibitem{DL}
Baader, F., Horrocks, I., Sattler, U.: Description logics as ontology languages
  for the semantic web. In: Mechanizing Mathematical Reasoning, pp. 228--248.
  Springer (2005)

\bibitem{brunoHAR}
Bruno, B., Grosinger, J., Mastrogiovanni, F., Pecora, F., Saffiotti, A.,
  Sathyakeerthy, S., Sgorbissa, A.: Multi-modal sensing for human activity
  recognition. In: Proceedings of 24th IEEE International Symposium on Robot
  and Human Interactive Communication (RO-MAN). pp. 594--600. Kobe, Japan
  (2015)

\bibitem{luca}
Buoncompagni, L., Bruno, B., Giuni, A., Mastrogiovanni, F., Zaccaria, R.:
  Arianna: towards a new paradigm for assistive technology at home. In:
  Proceedings of 8th Italian Forum on Ambient Assisted Living - ForItAAL.
  Genova, Italy (2015)

\bibitem{luca2017forgettings}
Buoncompagni, L., Dinale, A., Mastrogiovanni, F.: The importance of remembering
  and forgetting in smart agents. In: Proceedings of 15th International
  Conference on Ubiquitous Robots and Ambient Intelligence (URAI). Hawaii, USA
  (2018)

\bibitem{chen}
Chen, L., Nugent, C.D., Wang, H.: A knowledge-driven approach to activity
  recognition in smart homes. IEEE Transactions on Knowledge and Data
  Engineering  \textbf{24}(6),  961--974 (2012)

\bibitem{hp}
Cohen, J.E.: Human population: the next half century. In: Science, vol.~302,
  pp. 1172--1175. American Association for the Advancement of Science (2003)

\bibitem{generative1}
Cook, D.J., Krishnan, N.C., Rashidi, P.: Activity discovery and activity
  recognition: A new partnership. IEEE Transactions on Cybernetics
  \textbf{43}(3),  820--828 (2013)

\bibitem{deduction1994}
Donini, F.M., Lenzerini, M., Nardi, D., Schaerf, A.: Deduction in concept
  languages: From subsumption to instance checking. Journal of Logic and
  Computation  \textbf{4}(4),  423--452 (1994)

\bibitem{swrl}
Horrocks, I., Patel-Schneider, P.F., Boley, H., Tabet, S., Grosof, B., Dean,
  M., et~al.: Swrl: A semantic web rule language combining owl and ruleml. W3C
  Member Submission  \textbf{21}, ~79 (2004)

\bibitem{Liu16}
Liu, L., Stroulia, E., Nikolaidis, I., Miguel-Cruz, A., Rincon, A.R.: Smart
  homes and home health monitoring technologies for older adults: A systematic
  review. International Journal of Medical Informatics  \textbf{91},  44--59
  (2016)

\bibitem{survey2018}
Mshali, H., Lemlouma, T., Moloney, M., Magoni, D.: A survey on health
  monitoring systems for health smart homes. International Journal of
  Industrial Ergonomics  \textbf{66},  26--56 (2018)

\bibitem{riboni}
Riboni, D., Pareschi, L., Radaelli, L., Bettini, C.: Is ontology-based activity
  recognition really effective? In: Pervasive Computing and Communications
  Workshops (PERCOM Workshops), 2011 IEEE International Conference on. pp.
  427--431. IEEE (2011)

\bibitem{uoftoroaodl}
Salguero, A.G., Espinilla, M., Delatorre, P., Medina, J.: Using ontologies for
  the online recognition of activities of daily living. Sensors
  \textbf{18}(4), ~1202 (2018)

\bibitem{Scalmato}
Scalmato, A., Sgorbissa, A., Zaccaria, R.: Describing and recognizing patterns
  of events in smart environments with description logic. IEEE Transactions on
  Cybernetics  \textbf{43}(6),  1882--1897 (2013)

\bibitem{cameraUse}
Veeraraghavan, A., Roy-Chowdhury, A.K., Chellappa, R.: Matching shape sequences
  in video with applications in human movement analysis. IEEE Transactions on
  Pattern Analysis and Machine Intelligence  \textbf{27}(12),  1896--1909
  (2005)

\end{thebibliography}
\end{document}